\documentclass[10pt,twocolumn,letterpaper]{article}

\usepackage{iccv}
\usepackage{times}
\usepackage{epsfig}
\usepackage{graphicx}
\usepackage{amsmath}
\usepackage{amssymb}

\usepackage{multirow}
\usepackage{subfigure}
\usepackage{booktabs}

\usepackage[breaklinks=true,bookmarks=false]{hyperref}

\iccvfinalcopy 

\def\iccvPaperID{****} 

\ificcvfinal\fi

\begin{document}

\title{Gait Recognition via Effective Global-Local Feature Representation and Local Temporal Aggregation}


\author{Beibei Lin\textsuperscript{1}, Shunli Zhang\textsuperscript{1}\thanks{Shunli Zhang is the corresponding author.} ~and Xin Yu\textsuperscript{2} \\
\textsuperscript{1} Beijing Jiaotong University, \\ \textsuperscript{2} University of Technology Sydney\\
{\tt\small \{18126289, slzhang\}@bjtu.edu.cn, xin.yu@uts.edu.au}}


\maketitle
\ificcvfinal\thispagestyle{empty}\fi

\begin{abstract}
Gait recognition is one of the most important biometric technologies and has been applied in many fields. Recent gait recognition frameworks represent each gait frame by descriptors extracted from either global appearances or local regions of humans. However, the representations based on global information often neglect the details of the gait frame, while local region based descriptors cannot capture the relations among neighboring regions, thus reducing their discriminativeness. In this paper, we propose a novel feature extraction and fusion framework to achieve discriminative feature representations for gait recognition. Towards this goal, we take advantage of both global visual information and local region details and develop a Global and Local Feature Extractor (GLFE). Specifically, our GLFE module is composed of our newly designed multiple global and local convolutional layers (GLConv) to ensemble global and local features in a principle manner. Furthermore, we present a novel operation, namely Local Temporal Aggregation (LTA), to further preserve the spatial information by reducing the temporal resolution to obtain higher spatial resolution. With the help of our GLFE and LTA, our method significantly improves the discriminativeness of our visual features, thus improving the gait recognition performance. Extensive experiments demonstrate that our proposed method outperforms state-of-the-art gait recognition methods on two popular datasets. 
\end{abstract}


\section{Introduction}
Gait recognition is a biometric technology depending on the uniqueness of human walking posture. Since human gait can be captured in long-distance conditions and the recognition process does not need the subject's cooperation, gait recognition can be widely applied in many fields, such as video surveillance, intelligent transportation, etc. However, the performance of gait recognition suffers from many conditions, e.g. changing clothing, carrying conditions, cross-view, speed change, and resolution \cite{connor2018biometric,liao2017pose,yu2006framework}. Therefore, improving the performance of gait recognition in a complex external environment is still highly desirable.

Recently, many existing gait recognition methods employ convolutional neural networks (CNNs) to generate gait feature representations and achieve better recognition performance than the traditional approaches. In general, the feature representations can be divided into two categories: global and local feature based representation. 
Global feature based representation methods extract gait features from whole gait frames.
Shiraga et al.~\cite{shiraga2016geinet} use 2D CNNs to extract global gait features from Gait Energy Image (GEI). Chao et al. \cite{chao2019gaitset} also use 2D CNNs to extract global features at the frame-level.
Local feature based representation methods extract and combine local gait features from local gait parts.
Zhang et al.~\cite{zhang2019cross} partition human gaits into different local parts and use multiple separate 2D CNNs to extract local features. Fan et al.~\cite{fan2020gaitpart} design a focal convolution layer to further extract local features from feature maps.

However, the aforementioned methods only utilize either global or local feature for representation, thus limiting the recognition performance. In particular, the global feature representations may not pay enough attention to the details of the gait, while the local feature representations may lose the global context information of the gait and neglect the relations among local regions. Moreover, Wolf et al.\cite{wolf2016multi} introduce 3D CNNs to extract the robust spatial-temporal gait features. However, traditional 3D CNNs requires fixed-length gait sequences for classification and thus are not able to address different length of videos directly.

To address the above issues, in this paper, we propose a novel cross-view gait recognition framework by learning effective representations from global and local features. Specifically, we build a new feature extraction module, called Global and Local Feature Extractor(GLFE), in the 3D CNNs framework to attain discriminative representations from both global and local information of gait frames. 

In the GLFE module, we design a new Global and Local Convolutional layer (GLConv) to extract both global and local features in a principle way. The global feature extractor focuses on the entire visual gait appearance, while the local one pay attention to the gait details. Then, the GLFE module is composed of multiple GLConv layers.  
By combining global and local gait feature maps, the GLFE module is able to attain more discriminative feature representation.

Since existing 2D CNNs based methods \cite{fan2020gaitpart,chao2019gaitset} usually use a spatial pooling layer to downsample feature resolution, the spatial information will be lost gradually. To fully exploit spatial information, we develop a novel Local Temporal Aggregation (LTA) operation to replace the traditional spatial pooling layer and aggregate temporal information in local clips. In this fashion, we leverage the temporal resolution to attain higher spatial resolution.  

Since the proposed method employs 3D convolutions, the temporal convolution is applied to aggregating temporal information.

The main contributions of this paper are three-fold.

1) We present a novel gait recognition framework to obtain a discriminative gait representation. In this framework, we introduce a new Global and Local Feature Extraction (GLFE) module with our Global and Local Convolutional layers (GLConv).

2) We propose a novel Local Temporal Aggregation (LTA) operation to aggregate local temporal information while preserving the spatial information.

3) The proposed method has been evaluated on public datasets CASIA-B and OUMVLP. The experimental results demonstrate that it can achieve state-of-the-art performance, especially in complex conditions. 

\section{Related Work}
One of the typical gait recognition methods is to model 3D human  \cite{ariyanto2011model,bodor2009view,zhao20063d}. The 3D human model often uses multiple cameras to capture gait data from different view angles and reconstruct the human 3D model, and then is utilized to generate 3D gait features for recognition. 
Although the 3D human model can carry more information and has strong robustness in cross-view condition, it is difficult to build a 3D human gait model in real scenes. Meanwhile, building human 3D model requires complex computation. Hence, most recent researchers conduct human gait recognition based on 2D gait data, which makes the cross-view become the challenging factor.

Recently, in the works based on 2D gait data, gait sequences are usually synthesized into a gait template for recognition. To degrade the effect of cross-view, existing gait recognition methods in this fashion can be divided into two types.
The first type makes use of hand-crafted view-invariant feature \cite{goffredo2009self,liu2011joint,jean2009towards} which normalizes different view angles to a specific view angle. Although normalizing views can address the problem of cross-view, it causes the gait information loss during the transformation process.
The second type utilizes the View Transformation Model (VTM) \cite{kusakunniran2010support,xing2016complete} to build the correlation between two different view angles. Then, one gait view can be transformed to another for recognition during the test stage. However, it is difficult to fully collect all view angles of the human gait in reality.

\begin{figure*}[ht]
\centering
\includegraphics[width=1\textwidth]{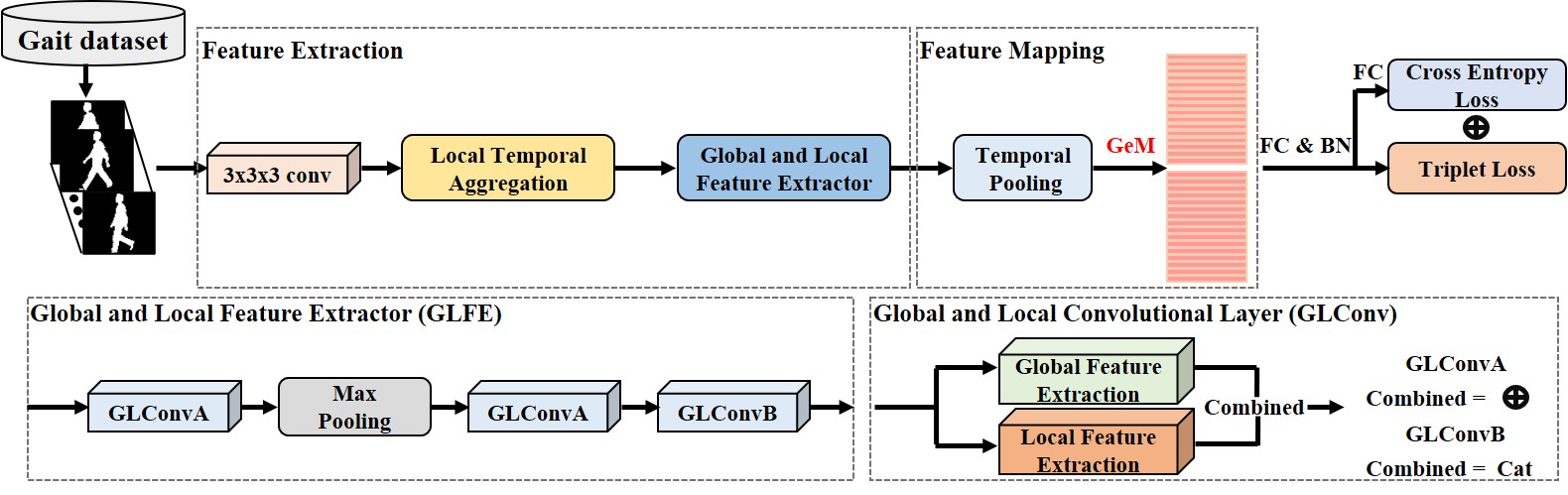}
\caption{Overview of the whole gait recognition framework.}
\vspace*{-1em}
\label{overview}
\end{figure*}

Inspired by the great achievements of CNNs in face recognition and action recognition, some researchers leverage CNNs for gait recognition. Although the cross-view is not addressed explicitly, CNNs can extract more robust gait features, improving the feature representation ability. 
Recently, designing gait recognition methods based on 2D or 3D convolutional networks becomes popular. Shiraga et al. \cite{shiraga2016geinet} propose the GEInet structure based on 2D convolution. They utilize two convolutional layers to extract features from GEI. Zhang et al.\cite{zhang2016siamese} introduce a Siamese framework with 2D convolution which also works on GEI. Wu et al.\cite{wu2016comprehensive} propose the LB and MT 2D CNNs method. However, the temporal information of gait sequences may be lost during the synthetic process of GEI. Therefore, some researchers first extract gait features at the frame-level gait image instead of GEI and then conduct operations to generate gait templates \cite{chao2019gaitset,zhang2019cross,fan2020gaitpart}. Chao et al.\cite{chao2019gaitset} use a statistical function to build gait templates after frame-level feature extraction. 
However, these methods only extract the feature from each frame, which uses spatial information but abandons temporal relation between the frames. 
Zhang et al.\cite{zhang2019cross} introduce a temporal attention mechanism to learn the attention scores of each frame, which are used to adaptively adjust the weights of specific frames. Fan et al.\cite{fan2020gaitpart} propose a novel framework named GaitPart, to extract spatial features from gait sequences and then models the temporal dependencies of the extracted 2D features.
To make use of both the spatial and temporal information, 3D convolution is introduced\cite{wolf2016multi,lin2020gait}. For example, Wolf et al. \cite{wolf2016multi} utilize 3D convolution for feature extraction in gait recognition. However, traditional 3D CNNs need to take a fixed-length gait sequence as an input, which cannot deal with different length videos adaptively.
In this paper, the proposed method not only uses 3D convolution but also introduces statistics function to aggregate temporal information for representation.

To further improve the feature representation ability, some  researchers \cite{zhang2019cross,fan2020gaitpart} utilize local gait features to replace global features. Zhang et al.\cite{zhang2010low} partition human gait image into four different parts as local human gaits. Then, they use multiple 2D CNNs to extract local gait feature of each local part. Fan et al. \cite{fan2020gaitpart} propose a novel convolution layer, called Focal Convolution Layer, to extract local gait features from local feature maps. Although it can learn more details, local gait features neglect the relation of different local regions.
Therefore, in this paper, we propose a novel GLFE module to extract more comprehensive features, which contain both global and local information. 

\section{Proposed Method} \label{method}
In this section, we first overview the framework of the proposed method. Then we describe the key components of the proposed method, including Local Temporal Aggregation (LTA), Global and Local Feature Extractor (GLFE) and Generalized-Mean (GeM) pooling layer \cite{radenovic2018fine}. Finally, the details of training and testing are presented.

\subsection{Overview}
The overview of the proposed method is shown in Fig.~\ref{overview}, which aims to extract more comprehensive feature representation for gait recognition, including three key components. First, we use a convolution to extract the shallow features from the original input sequence. Next, the Local Temporal Aggregation (LTA) operation is designed to aggregate the temporal information and preserve more spatial information for trade off. After that, the Global and Local Feature Extractor (GLFE) is implemented to extract the combined feature ensembling both global and local information. Then, we leverage temporal pooling and GeM pooling layer to implement feature mapping. Finally, we choose the triplet loss \cite{chao2019gaitset,fan2020gaitpart} and cross entropy loss to train the proposed model.

\subsection{Local Temporal Aggregation}
Previous works \cite{chao2019gaitset,fan2020gaitpart} use a specific pattern ``CL-SP-CL-SP-CL" to extract features, where CL means convolutional layer and SP denotes the spatial pooling layer. However, the spatial information may be lost due to the SP downsampling operation. Considering that temporal information in a gait sequence is periodic, we present the LTA operation to replace the first spatial pooling layer, which can integrate temporal information of local clips and maintain more spatial information. 

Assume that $X_{in}\in \mathbb{R}^{C_{1} \times T_{1} \times H_{1} \times W_{1}}$ is the input of local temporal aggregation, where $C_1$ is the number of channels, $T_{1}$ is the length of the gait sequence and ($H_{1}$,$W_{1}$) is the image size of each frame. The process can be formulated as follows
\begin{equation}
X_{LTA} =  f_{a\times a\times a}^{b \times1\times1}(X_{in}),
\end{equation}
where $f_{a\times a\times a}^{b \times1\times1}(\cdot)$ denotes the 3D convolution operation with kernel size $a$ and temporal stride $b$. $X_{LTA} \in \mathbb{R}^{C_{2} \times T_{2} \times H_{1} \times W_{1}}$ is the output of LTA operation.

\begin{figure*}[ht]
\centering
\includegraphics[width=1\textwidth]{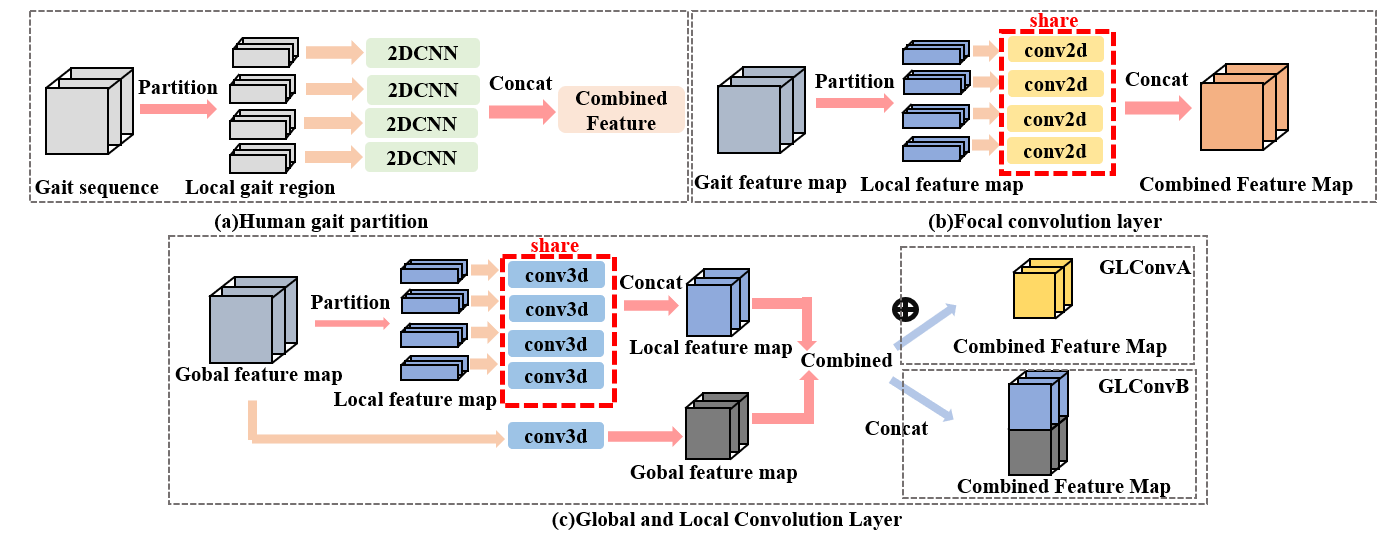}
\caption{Architectures of different gait feature extractors. 
$\oplus$ means element-wise addition, ``Concat'' means concatenating the feature maps of different parts horizontally, ``2DCNN'' denotes a 2D convolution neural network, ``conv2d'' denotes a 2D convolution operation and ``conv3d'' means a 3D convolution operation.
}
\vspace*{-1em}
\label{image_GLconv}
\end{figure*}

\subsection{Global and Local Feature Extractor}
\label{Sec_GLFE}
Except the global gait features \cite{chao2019gaitset,shiraga2016geinet}, some recent researchers propose different gait recognition frameworks which utilize local gait information, as shown in Fig.\ref{image_GLconv}(a)(b) \cite{zhang2019cross,fan2020gaitpart}.
For example, Zhang et al.\cite{zhang2019cross} propose the ACL framework to extract local gait features by using multiple separate 2D CNNs. Fan et al.\cite{fan2020gaitpart} develop a focal convolutional layer to extract the local features and then combine them. 
Although it contains more detailed information than the global gait features, the local gait features do not pay attention to the relations among local regions. 
Hereby, we propose a novel GLFE module to extract features, which can take advantage of both global and local information. The GLFE module is implemented by the GLConv layer, which contains global and local feature extractors. The global feature extractor can extract the whole gait information, while the local feature extractor is used to extract more details from local feature maps. The GLConv has two different structures due to different combinations, e.g. GLConvA and GLConvB. The GLFE module includes four layers, ``GLConvA-SP-GLConvA-GLConvB” as shown in Fig.\ref{overview}. 

The GLConv layer is shown in Fig.\ref{image_GLconv}(c). Assume that its input is $X_{global} \in \mathbb{R}^{c_{1} \times t \times h \times w}$, where $c_{1}$ is the number of channels, $t$ is the length of feature maps and ($h$,$w$) is the image size of each frame. We first partition the global feature map into \textit{n}-parts as local feature maps $X_{local} = \left \{ X_{local}^{i} | i=1,...,n \right \}$, where $n$ is the number of partitions and $ X_{local}^{i}  \in \mathbb{R}^{c_{1} \times t \times \frac{h}{n} \times w}$ corresponds to the \textit{i}-th local gait part. Then, we use 3D convolutions to extract global and local gait features, respectively. Note that all local feature maps share the same convolutional weights. 
There are two ways to combine the global and local feature maps, i.e. by element-wise addition (GLconvA) or by concatenation (GLconvB). The GLconvA and GLconvB layers can be representated as
\begin{equation}
Y_{GLConvA} = Y_{global} + Y_{local} \in \mathbb{R}^{c_{2} \times t \times h\times w}
\end{equation}
and
\begin{equation}
Y_{GLConvB} = cat
\begin{Bmatrix}
Y_{global}
\\ 
Y_{local}
\end{Bmatrix}\in \mathbb{R}^{c_{2} \times t \times 2h\times w},
\end{equation}
where $cat$ means concatenating operation. $Y_{global}$ and $Y_{local}$  can be represented as
\begin{equation}
Y_{global} = f_{3\times3\times3}(X_{global})\in \mathbb{R}^{c_{2} \times t \times h\times w}
\end{equation}
and
\begin{equation}
Y_{local} = cat\begin{Bmatrix}
f_{3\times3\times3}^{'}(Y_{local}^{1})\\
f_{3\times3\times3}^{'}(Y_{local}^{2})\\
...\\
f_{3\times3\times3}^{'}(Y_{local}^{n})
\end{Bmatrix} \in \mathbb{R}^{c_{2} \times t \times h\times w},
\end{equation}
where $f_{3\times 3\times 3}(\cdot)$ and $f_{3\times 3\times 3}^{'}(\cdot)$ denote 3D convolutions with kernel size 3.

Based on the above two forms of GLGonv layer, the GLFE module can be built to extract gait features after LTA operation. 
Experimentally, GLConvA is used to implement the first few GLGonv blocks and GLConvB is ulitized to realize the last one in the GLFE module.

\subsection{Feature Mapping}
\label{Sec_FP}
Since the length of the input gait sequences may be different, we introduce temporal pooling to aggregate the temporal information of the whole sequence \cite{chao2019gaitset,fan2020gaitpart}. Assume that $X_{GLFE}\in \mathbb{R}^{C_{3} \times T_{2} \times H_{2} \times W_{2}}$ is the output of GLFE module, where $C_3$ is the number of input channels, $T_{2}$ is the length of feature maps and ($H_{2}$,$W_{2}$) is the spatial size of the feature in each frame. Because of the spatial pooling layer in the GLFE module, the spatial size becomes ($H_{2}$,$W_{2}$), while the length of feature maps remains unchanged. The temporal pooling $TP(\cdot)$ can be realized by
\begin{equation} \label{temporal_agg_equ}
  Y_{TP} =  F_{Max}^{T_{2}\times1\times1}(X_{GLFE}),
\end{equation}
where $F_{Max}^{T_{2}\times1\times1}(\cdot)$ means Max-pooling layer. $Y_{TP} \in \mathbb{R}^{C_{3} \times 1 \times H_{2} \times W_{2}}$ is the output of temporal pooling.

To improve feature representation ability, researchers develop the spatial feature mapping operation with weighted sum \cite{chao2019gaitset,fan2020gaitpart}. After temporal pooling, gait feature maps are split into strips and two statistical functions, max and average, are used to aggregate each strip's information. The spatial feature mapping can be represented as
\begin{equation}
  Y_{MA} = \alpha F_{Max}^{1\times1\times W_{2}}(Y_{TP}) + \beta F_{Avg}^{1\times1\times W_{2}}(Y_{TP}),
  \label{Equ_MA}
\end{equation}
where $Y_{MA} \in \mathbb{R}^{C_{3} \times 1 \times H_{2} \times 1}$ is the output of spatial feature mapping. However, the weighted sum strategy is inflexible because trade-off parameters are predefined manully. 

Hereby, we introduce the Generalized-Mean pooling (GeM) to integrate the spatial information adaptively. The GeM pooling layer $GeM(\cdot)$ can be represented as
\begin{equation}
  Y_{GeM} = (F_{Avg}^{1\times1\times W_{2}}( (Y_{TP})^{p}))^{\frac{1}{p}},
  \label{Equ_GeM}
\end{equation}
where $Y_{GeM} \in \mathbb{R}^{C_{3} \times 1 \times H_{2} \times 1}$ is the output of GeM operation. $p$ is the parameter which can be learned by network training. Specifically, if $p=1$, $Y_{GeM}$ is equal to $Avg^{1\times1\times W_{2}}(Y_{TP})$, and if $p \rightarrow \infty$, $Y_{GeM}$  is equal to $Max^{1\times1\times W_{2}}(Y_{TP})$. 
Then, we use multiple separate fully connected layers to further aggregate the information from channels of $Y_{GeM}$. The feature mapping can be defined as
\begin{equation}
Y_{out} =  {F}_{sfc}(Y_{GeM})\in \mathbb{R}^{C_{4} \times 1 \times H_{2} \times 1},
\end{equation}
$Y_{out}$ is the output of feature mapping with $H_{2}$ horizontal features, each of which has $C_{4}$ channels. 
$F_{sfc}(\cdot)$ denotes the multiple separate fully connected layers.

\subsection{Loss Function}\label{loss_function}
To effectively train the proposed gait recognition model, we simultaneously utilize the triplet loss \cite{hermans2017defense} and cross entropy loss . 

The triplet loss can improve the inter-class distance and reduce the intra-class distance, which can help the cross entropy loss to identify Human IDs. 
In the training stage, each horizontal feature of $Y_{out}$ is fed into the combined loss function to calculate the loss independently. The combined loss function $L_{combined}$ can be defined as
\begin{equation}
L_{combined} = L_{tri} + L_{cse},
\end{equation}
where $L_{tri}$ and $L_{cse}$ mean the triplet loss and cross entropy loss, respectively. $L_{tri}$ can be defined as
\begin{equation}
\label{tri}
L_{tri} = [D(F(i),F(k)) - D(F(i),F(j)) + m]_{+},
\end{equation}
where $i$ and $j$ are the samples from the same class A, while $k$ represents the sample from class B. $F(\cdot)$ denotes the feature extraction and mapping operation of the proposed method. $D(d_1, d_2)$ is the Euclidean distance between $d_1$ and $d_2$. $m$ is the margin of the triplet loss. The operation $[\gamma]_{+}$ is equal to $max(\gamma,0)$.

\subsection{Training Details and Test}
\label{Traintest}
\noindent \textbf{Training Stage.} 
During the training stage, we first feed the input gait sequences into the proposed network to generate gait feature representation $Y_{out}$. Then, the combined loss function is used to compute the loss and Batch ALL (BA) is adopted as the sampling strategy, which is the same as \cite{hermans2017defense,chao2019gaitset,fan2020gaitpart}. Specifically, each batch contains $P$ subject IDs, and $K$ samples are selected from each subject ID. Correspondingly, the batch size is $P \times K$. During the training stage, considering the memory limit, we set the length of input sequence to T. 

\noindent \textbf{Test Stage.} 
During the test stage, the whole gait sequences are put into the proposed network to obtain gait features $Y_{out}$. Then, the $Y_{out} \in \mathbb{R}^{C_{4} \times 1 \times H_{2} \times 1}$ can be flattened to a feature vector with dimension $C_{4} \times H_{2}$ and then taken as a sample. To calculate Rank-1 accuracy, the test dataset is divided into two sets, i.e. the gallery set and the probe set. The gallery set is regarded as the standard view to be retrieved, while the feature vectors of the probe are used to match the feature vectors from the gallery view. Multiple metric strategies, such as Euclidean distance and cosine distance, can be used to calculate similarity  between the samples from gallery and probe set. Specifically, Euclidean distance is selected as the metric strategy.

\section{Experiments}

\subsection{Datasets}

We evaluate the performance of the proposed method on two commonly used datasets, i.e. CASIA-B \cite{yu2006framework} and OUMVLP \cite{takemura2018multi}  datasets.

\noindent \textbf{CASIA-B.} The CASIA-B dataset \cite{yu2006framework} is the largest cross-view gait database. It includes 124 subjects, each of which has 10 groups of videos. Among these groups, six of them are sampled in normal walking (NM), two groups are in walking with a bag (BG), and the rest are in walking in coats (CL). Each group contains 11 gait sequences from different angles ( $0^\circ$-$180^\circ$ and the sampling interval is $18^\circ$). Therefore, there are 124(subject) $\times$ 10(groups) $\times$ 11(view angle) = 13,640 gait sequences in CASIA-B. The gait sequence of each subject are divided into training set and test set. During the training stage, three training settings are configured according to the different training scales \cite{chao2019gaitset}: small-scale training (ST), medium-scale training (MT), and large-scale training (LT). For these three settings, 24, 62, and 74 subjects are chosen as the training set and the rest 100, 62, and 50 subjects are used for test, respectively. All gait data from the training set are used to train the proposed model during the training stage. In the test stage, the sequences NM\#01-NM\#04 are taken as the gallery set, while the sequences NM\#05-NM\#06, BG\#01-BG\#02, and CL\#01-CL\#02 are considered as the probe set to evaluate the performance.

\noindent \textbf{OUMVLP.} The OUMVLP dataset \cite{takemura2018multi} is one of the largest gait recognition databases, which contains 10,307 subjects in total. Each subject contains two groups of videos, Seq\#00 and Seq\#01. Each group of sequences are captured from 14 angles: $0^\circ$-$90^\circ$ and $180^\circ$-$270^\circ$ with sampling interval $15^\circ$. We adopt the same protocol (5,153 subjects are taken as training data and 5,154 subjects are used as test data) as \cite{chao2019gaitset} and \cite{fan2020gaitpart} to evaluate the performance of the proposed method. In the test stage, the sequences in Seq\#01 are taken as the gallery set, while the sequences in Seq\#00 are regarded as the probe set for evaluation.

\begin{table*}[htbp]
  \centering
  \vspace*{-1em}
  \caption{Network structures of the proposed method.}
  \resizebox{0.95\textwidth}{!}{
    \begin{tabular}{c|c|c|c|c|c|c|c|c|c|c|c}
    \toprule
    \multicolumn{6}{c|}{CASIA-B}                 & \multicolumn{6}{c}{OUMVLP} \\
    \hline
    Layer Name & In\_C & Out\_C & Kernel & Global & N-part & Layer Name & In\_C & Out\_C & Kernel & Global & N-part \\
    \midrule
    \multirow{2}[2]{*}{Conv3d} & \multirow{2}[2]{*}{\checkmark} & \multirow{2}[2]{*}{32} & \multirow{2}[2]{*}{(3, 3, 3)} & \multirow{2}[2]{*}{\checkmark} & \multirow{2}[2]{*}{$\times$} & Conv3d & 1     & 32    & (3, 3, 3) & \checkmark     & $\times$ \\
\cline{7-12}          &       &       &       &       &       & Conv3d & 32    & 32    & (3, 3, 3) & \checkmark     & $\times$ \\
    \hline
    LTA   & 32    & 32    & (3, 1, 1) & $-$     & $-$     & LTA   & 32    & 32    & (3, 1, 1) & $-$     & $-$ \\
    \hline
    \multirow{2}[2]{*}{GLConvA} & \multirow{2}[2]{*}{32} & \multirow{2}[2]{*}{32} & \multirow{2}[2]{*}{(3, 3, 3)} & \multirow{2}[2]{*}{\checkmark} & \multirow{2}[2]{*}{8} & GLConvA & 32    & 64    & (3, 3, 3) & \checkmark     & 8 \\
\cline{7-12}          &       &       &       &       &       & GLConvA & 64    & 64    & (3, 3, 3) & \checkmark     & 8 \\
    \hline
    Max Pooling & $-$     & $-$     & (1, 2, 2) & $-$     & $-$     & Max Pooling & $-$     & $-$     & (1, 2, 2) & $-$     & $-$ \\
    \hline
    \multirow{2}[2]{*}{GLConvA} & \multirow{2}[2]{*}{64} & \multirow{2}[2]{*}{128} & \multirow{2}[2]{*}{(3, 3, 3)} & \multirow{2}[2]{*}{\checkmark} & \multirow{2}[2]{*}{8} & GLConvA & 64    & 128   & (3, 3, 3) & \checkmark     & 8 \\
\cline{7-12}          &       &       &       &       &       & GLConvA & 128   & 128   & (3, 3, 3) & \checkmark     & 8 \\
    \hline
    \multirow{2}[2]{*}{GLConvB} & \multirow{2}[2]{*}{128} & \multirow{2}[2]{*}{128} & \multirow{2}[2]{*}{(3, 3, 3)} & \multirow{2}[2]{*}{\checkmark} & \multirow{2}[2]{*}{8} & GLConvA & 128   & 256   & (3, 3, 3) & \checkmark     & 8 \\
\cline{7-12}          &       &       &       &       &       & GLConvB & 256   & 256   & (3, 3, 3) & \checkmark     & 8 \\
    \bottomrule
    \end{tabular}%
    }
  \label{network}%
  \vspace*{-1em}
\end{table*}%

\subsection{Implementation Details}
We adopt the same preprocessing approach as \cite{chao2019gaitset} to obtain gait silhouettes for CASIA-B and OUMVLP datasets. The image of each frame is normalized to the size $64 \times 44$. 
The network parameters are shown in Table.\ref{network}. 
$m$ in Equ.\ref{tri} is set to 0.2. $p$ in Equ.\ref{Equ_GeM} is initialized to 6.5. The batch size parameters P and K are both set to 8 in CASIA-B dataset. Since OUMVLP dataset is much larger than CASIA-B, the batch size P$\times$ K is set to 32$\times$8 = 256. 
During the training stage, the length of input gait sequences is set to 30. During the test stage, the whole gait sequences are put into the proposed model to extract gait features. In the setting of ST, MT and LT, the epoch number is set to 60K, 80K and 80K, receptively. All experiments take Adam as the optimizer. The weight decay is set to 5e-4 for CASIA-B dataset.  For the setting of ST, the learning rate is set to 1e-4. For the settings of MT and LT, the learning rate is first set to 1e-4, and reset to 1e-5 after 70K iterations.
For the OUMVLP dataset, the epoch number is set to 210K. The learning rate is first set to 1e-4, and reset to 1e-5 and 5e-6 after 150K and 200K iterations, respectively. In particular, since OUMVLP has 100 times more IDs than the CASIA dataset, we add the label smooth operation into the cross entropy loss \cite{hou2020gait}. The weight decay is first set to 0 and reset to 5e-4 after 200K for OUMVLP dataset.


\begin{table*}[ht]
  \scriptsize
  \centering
  \vspace*{-1em}
   \caption{Rank-1 accuracy (\%) on CASIA-B under all view angles, different settings and conditions, excluding identical-view case.
  }
  \resizebox{0.98\textwidth}{!}{
    \begin{tabular}{c|c|c|c|c|c|c|c|c|c|c|c|c|c|c}
    \toprule
    \multicolumn{3}{c|}{Gallery NM\#1-4}  &\multicolumn{12}{c}{$0^{\circ}$-$180^{\circ}$} \\
    \hline
    \multicolumn{3}{c|}{Probe}    & $0^{\circ}$     & $18^{\circ}$    & $36^{\circ}$    & $54^{\circ}$    & $72^{\circ}$    & $90^{\circ}$    & $108^{\circ}$   & $126^{\circ}$   & $144^{\circ}$   & $162^{\circ}$   & $180^{\circ}$  & Mean\\
    \midrule

    \multicolumn{1}{c|}{\multirow{9}[2]{*}{\textbf{ST(24)}}} & \multicolumn{1}{c|}{\multirow{5}[2]{*}{NM\#5-6}} 
    & ViDP &  $-$     &  $-$     &   $-$    & 59.1  &   $-$    & 50.2  &   $-$    & 57.5  &    $-$   &   $-$    &  $-$     & $-$ \\
&       & CMCC & 46.3  &   $-$    &   $-$    & 52.4  &    $-$   & 48.3  &     $-$  & 56.9  &    $-$   &    $-$   &      $-$ &  $-$\\
&       & CNN-LB & 54.8  &   $-$    &    $-$   & 77.8  &   $-$    & 64.9  &  $-$     & 76.1  &  $-$     &    $-$   &    $-$   & $-$ \\
&       & GaitSet & 64.6  & 83.3  & 90.4 & 86.5  & 80.2  & 75.5 & 80.3  & 86.0  & 87.1  & 81.4  & 59.6  & 79.5  \\
&       & Ours  & \textbf{77.0} & \textbf{87.8} & \textbf{93.9} & \textbf{92.7} & \textbf{83.9} & \textbf{78.7} & \textbf{84.7} & \textbf{91.5} & \textbf{92.5} & \textbf{89.3} & \textbf{74.4} & \textbf{86.0} \\

\cline{2-15}          & \multicolumn{1}{c|}{\multirow{2}[2]{*}{BG\#1-2}} & GaitSet & 55.8  & 70.5  & 76.9  & 75.5  & 69.7  & 63.4  & 68.0  & 75.8  & 76.2  & 70.7  & 52.5  & 68.6  \\
&       & Ours  & \textbf{68.1} & \textbf{81.2} & \textbf{87.7} & \textbf{84.9} & \textbf{76.3} & \textbf{70.5} & \textbf{76.1} & \textbf{84.5} & \textbf{87.0} & \textbf{83.6} & \textbf{65.0} & \textbf{78.6} \\

\cline{2-15}          & \multicolumn{1}{c|}{\multirow{2}[2]{*}{CL\#1-2}} & GaitSet & 29.4  & 43.1  & 49.5  & 48.7  & 42.3  & 40.3  & 44.9  & 47.4  & 43.0  & 35.7  & 25.6  & 40.9  \\
    %
&       & Ours  & \textbf{46.9} & \textbf{58.7} & \textbf{66.6} & \textbf{65.4} & \textbf{58.3} & \textbf{54.1} & \textbf{59.5} & \textbf{62.7} & \textbf{61.3} & \textbf{57.1} & \textbf{40.6} & \textbf{57.4} \\
\hline


    \multicolumn{1}{c|}{\multirow{12}[2]{*}{\textbf{MT(62)}}} & \multicolumn{1}{c|}{\multirow{4}[2]{*}{NM\#5-6}} & AE    & 49.3  & 61.5  & 64.4  & 63.6  & 63.7  & 58.1  & 59.9  & 66.5  & 64.8  & 56.9  & 44.0  & 59.3  \\
          &       & MGAN  & 54.9  & 65.9  & 72.1  & 74.8  & 71.1  & 65.7  & 70.0  & 75.6  & 76.2  & 68.6  & 53.8  & 68.1  \\
          &       & GaitSet & 86.8  & 95.2  & 98.0  & 94.5  & 91.5  & 89.1  & 91.1  & 95.0  & 97.4  & 93.7  & 80.2  & 92.0  \\
&       & Ours  & \textbf{93.9} & \textbf{97.6} & \textbf{98.8} & \textbf{97.3} & \textbf{95.2} & \textbf{92.7} & \textbf{95.6} & \textbf{98.1} & \textbf{98.5} & \textbf{96.5} & \textbf{91.2} & \textbf{95.9} \\

\cline{2-15}          & \multicolumn{1}{c|}{\multirow{4}[2]{*}{BG\#1-2}} & AE    & 29.8  & 37.7  & 39.2  & 40.5  & 43.8  & 37.5  & 43.0  & 42.7  & 36.3  & 30.6  & 28.5  & 37.2  \\
          &       & MGAN  & 48.5  & 58.5  & 59.7  & 58.0  & 53.7  & 49.8  & 54.0  & 51.3  & 59.5  & 55.9  & 43.1  & 54.7  \\
         &       & GaitSet & 79.9  & 89.8  & 91.2  & 86.7  & 81.6  & 76.7  & 81.0  & 88.2  & 90.3  & 88.5  & 73.0  & 84.3  \\
&       & Ours  & \textbf{88.5} & \textbf{95.1} & \textbf{95.9} & \textbf{94.2} & \textbf{91.5} & \textbf{85.4} & \textbf{89.0} & \textbf{95.4} & \textbf{97.4} & \textbf{94.3} & \textbf{86.3} & \textbf{92.1} \\

\cline{2-15}          & \multicolumn{1}{c|}{\multirow{4}[2]{*}{CL\#1-2}} & AE    & 18.7  & 21.0  & 25.0  & 25.1  & 25.0  & 26.3  & 28.7  & 30.0  & 23.6  & 23.4  & 19.0  & 24.2  \\
          &       & MGAN  & 23.1  & 34.5  & 36.3  & 33.3  & 32.9  & 32.7  & 34.2  & 37.6  & 33.7  & 26.7  & 21.0  & 31.5  \\
         &       & GaitSet & 52.0  & 66.0  & 72.8  & 69.3  & 63.1  & 61.2  & 63.5  & 66.5  & 67.5  & 60.0  & 45.9  & 62.5  \\
&       & Ours  & \textbf{70.7} & \textbf{83.2} & \textbf{87.1} & \textbf{84.7} & \textbf{78.2} & \textbf{71.3} & \textbf{78.0} & \textbf{83.7} & \textbf{83.6} & \textbf{77.1} & \textbf{63.1} & \textbf{78.3} \\

    \hline

    \multicolumn{1}{c|}{\multirow{14}[2]{*}{\textbf{LT(74)}}} & \multicolumn{1}{c|}{\multirow{6}[2]{*}{NM\#5-6}} & CNN-3D & 87.1  & 93.2  & 97.0  & 94.6  & 90.2  & 88.3  & 91.1  & 93.8  & 96.5  & 96.0  & 85.7  & 92.1  \\
          &       & CNN-Ensemble & 88.7  & 95.1  & 98.2  & 96.4  & 94.1  & 91.5  & 93.9  & 97.5  & 98.4  & 95.8  & 85.6  & 94.1  \\
          &       & GaitSet & 90.8  & 97.9 & 99.4 & 96.9  & 93.6  & 91.7  & 95.0  & 97.8  & 98.9 & 96.8  & 85.8  & 95.0  \\
&       & ACL   & 92.0  & 98.5  & \textbf{100.0} & \textbf{98.9} & 95.7  & 91.5  & 94.5  & 97.7  & 98.4  & 96.7  & 91.9  & 96.0  \\
&       & GaitPart & 94.1  & \textbf{98.6} & 99.3  & 98.5  & 94.0  & 92.3  & 95.9  & 98.4  & 99.2 & 97.8  & 90.4  & 96.2  \\
&       & Ours  & \textbf{96.0} & 98.3  & 99.0  & 97.9  & \textbf{96.9} & \textbf{95.4} & \textbf{97.0} & \textbf{98.9} & \textbf{99.3} & \textbf{98.8} & \textbf{94.0} & \textbf{97.4} \\

\cline{2-15}
          & \multicolumn{1}{c|}{\multirow{3}[2]{*}{BG\#1-2}} & CNN-LB & 64.2  & 80.6  & 82.7  & 76.9  & 64.8  & 63.1  & 68.0  & 76.9  & 82.2  & 75.4  & 61.3  & 72.4  \\
          &       & GaitSet & 83.8  & 91.2  & 91.8  & 88.8  & 83.3  & 81.0  & 84.1  & 90.0  & 92.2  & 94.4  & 79.0  & 87.2  \\
&       & GaitPart & 89.1  & 94.8 & 96.7 & 95.1 & 88.3  & 84.9 & 89.0  & 93.5  & 96.1  & 93.8  & 85.8  & 91.5  \\
&       & Ours  & \textbf{92.6} & \textbf{96.6} & \textbf{96.8} & \textbf{95.5} & \textbf{93.5} & \textbf{89.3} & \textbf{92.2} & \textbf{96.5} & \textbf{98.2} & \textbf{96.9} & \textbf{91.5} & \textbf{94.5} \\

\cline{2-15}          & \multicolumn{1}{c|}{\multirow{3}[2]{*}{CL\#1-2}} & CNN-LB & 37.7  & 57.2  & 66.6  & 61.1  & 55.2  & 54.6  & 55.2  & 59.1  & 58.9  & 48.8  & 39.4  & 54.0  \\
          &       & GaitSet & 61.4  & 75.4  & 80.7  & 77.3  & 72.1  & 70.1  & 71.5  & 73.5  & 73.5  & 68.4  & 50.0  & 70.4  \\
&       & GaitPart & 70.7  & 85.5  & 86.9  & 83.3  & 77.1  & 72.5  & 76.9  & 82.2  & 83.8  & 80.2  & 66.5  & 78.7  \\
&       & Ours  & \textbf{76.6} & \textbf{90.0} & \textbf{90.3} & \textbf{87.1} & \textbf{84.5} & \textbf{79.0} & \textbf{84.1} & \textbf{87.0} & \textbf{87.3} & \textbf{84.4} & \textbf{69.5} & \textbf{83.6} \\

    \bottomrule

    \end{tabular}%
    }

\label{comparision_casia}
\vspace*{-1em}
\end{table*}%

\subsection{Comparison with State-of-the-Art Methods}

\noindent \textbf{Evaluation on CASIA-B.} We compare the proposed method with the latest gait recognition methods, including ViDP~\cite{hu2013view}, CMCC~\cite{kusakunniran2013recognizing}, GaitSet~\cite{chao2019gaitset}, AE~\cite{yu2017invariant}, MGAN~\cite{he2018multi}, CNN-LB, CNN-3D, CNN-Ensemble~\cite{wu2016comprehensive}, ACL~\cite{zhang2019cross} and GaitPart~\cite{fan2020gaitpart} on CASIA-B. The experimental results are shown in Table.\ref{comparision_casia}. It can be observed that the proposed method achieves the best recognition accuracy at almost all angles. We further analyze the comparison results in details by considering different conditions (NM, BG, and CL) and different dataset scales.

We first explore the effect of different conditions (NM, BG, and CL). From Table. \ref{comparision_casia} we can observe that the accuracy has significantly decreased when the external environment changes. For example, the recognition accuracy of GaitPart in three conditions NM, BG, and CL is 96.2\%, 91.5\%, and 78.7\% with the setting of LT. For the proposed method, the recognition accuracy in these conditions is 97.4\%, 94.5\%, and 83.6\%, which outperforms GaitPart by 1.2\%, 3.0\%, and 4.9\%, respectively. The experimental results show that the proposed method has significant advantages in the BG and CL conditions, indicating that the proposed model can extract more robust gait features. Even in the ST and MT setting, we can see the similar recognition results that the proposed method also obtains the best performance. On the other hand, human gait can be collected from arbitrary view angles and conditions in a real scenario. Hereby, we  focus on the robustness of gait recognition under various external environment factors. Based on Table. \ref{comparision_casia}, we further calculate the average accuracy of three conditions. Table.\ref{Accuracy_mean} indicates that the comparison results of the proposed method with state-of-the-art gait recognition methods, including GaitSet~\cite{chao2019gaitset} and GaitPart~\cite{fan2020gaitpart}. It can be observed that the average accuracy of the proposed method is 91.8\%, which outperforms GaitSet and GaitPart by 7.6\% and 3.0\%, respectively.

We also discuss the comparison results with different setting of dataset scales. 
Chao et al. \cite{chao2019gaitset} provide an evaluation protocol, in which the performance of the method is evaluated on three different scales, i.e. ST, MT, and LT on CASIA-B. 
In this paper, we present the complete experimental results with these three settings. The experimental results are shown in Table. \ref{comparision_casia}. We can observe that the recognition accuracy of the GaitSet \cite{chao2019gaitset} is 79.5\%, 92.0\%, and 95.0\% for three settings under the condition of NM, respectively. For the proposed method, the corresponding recognition accuracy is 86.0\%, 95.9\%, and 97.4\%, which increases by 6.5\%, 3.9\%, and 2.4\%, respectively. Specifically, the proposed method can achieve larger improvement in the setting of small-scale dataset.


\begin{table}[htbp]
  \centering
  \caption{Comparison with GaitSet \cite{chao2019gaitset} and  GaitPart \cite{fan2020gaitpart} under the conditions of NM, BG and CL. }
    \begin{tabular}{c|c|c|c|c}
    \toprule
    Method & NM    & BG    & CL    & Mean \\
    \midrule
    GaitSet \cite{chao2019gaitset} & 95.0  & 87.2  & 70.4  & 84.2  \\
    \hline
    GaitPart \cite{fan2020gaitpart} & 96.2  & 91.5  & 78.7  & 88.8  \\
    \hline
    Ours  & \textbf{97.4} & \textbf{94.5} & \textbf{83.6} & \textbf{91.8}  \\
    \bottomrule
    \end{tabular}%
  \vspace*{-1em}
  \label{Accuracy_mean}%
\end{table}%

\begin{table*}[htbp]
  \centering
  \caption{Rank-1 accuracy (\%) on OUMVLP under 14 probe views excluding identical-view cases.}
  \resizebox{0.98\textwidth}{!}{
    \begin{tabular}{c|c|c|c|c|c|c|c|c|c|c|c|c|c|c|c}
    \toprule
    \multirow{2}[2]{*}{\textbf{Method}} & \multicolumn{14}{c|}{\textbf{Probe View}}                                                            & \multicolumn{1}{c}{\multirow{2}[2]{*}{\textbf{Mean}}} \\
\cline{2-15}    \multicolumn{1}{c|}{} & $0^{\circ}$  & $15^{\circ}$  & $30^{\circ}$  & $45^{\circ}$  & $60^{\circ}$  & $75^{\circ}$  & $90^{\circ}$ & $180^{\circ}$  & $195^{\circ}$  & $210^{\circ}$  & $225^{\circ}$  & $240^{\circ}$  & $255^{\circ}$  & $270^{\circ}$  &  \\
    \midrule
    GEINet & 23.2  & 38.1  & 48.0  & 51.8  & 47.5  & 48.1  & 43.8  & 27.3  & 37.9  & 46.8  & 49.9  & 45.9  & 45.7  & 41.0  & 42.5  \\
    \hline
    GaitSet & 79.3  & 87.9  & 90.0  & 90.1  & 88.0  & 88.7  & 87.7  & 81.8  & 86.5  & 89.0  & 89.2  & 87.2  & 87.6  & 86.2  & 87.1  \\
    \hline
    GaitPart & 82.6  & 88.9  & 90.8  & 91.0  & 89.7  & 89.9  & 89.5  & 85.2  & 88.1  & 90.0  & 90.1  & 89.0  & 89.1  & 88.2  & 88.7  \\
    \hline
    GLN   & 83.8  & 90.0  & 91.0  & 91.2  & 90.3  & 90.0  & 89.4  & 85.3  & \textbf{89.1} & \textbf{90.5} & \textbf{90.6} & 89.6  & 89.3  & 88.5  & 89.2  \\
    \hline
    Ours  & \textbf{84.9} & \textbf{90.2} & \textbf{91.1} & \textbf{91.5} & \textbf{91.1} & \textbf{90.8} & \textbf{90.3} & \textbf{88.5} & 88.6  & 90.3  & 90.4  & \textbf{89.6} & \textbf{89.5} & \textbf{88.8} & \textbf{89.7} \\
    \bottomrule
    \end{tabular}%
    }
    \vspace*{-1em}
  \label{comparision_oumvlp}%
\end{table*}%

\noindent \textbf{Evaluation on OUMVLP.} 
We further evaluate the performance of the proposed method on the OUMVLP dataset. For fair comparison, we take the same training and test protocols as the GaitSet and GaitPart methods\cite{chao2019gaitset,fan2020gaitpart}. 10,307 subjects are divided into two groups, where 5,153 subjects are used for training and the rest subjects are used for test. During the test stage, Seq\#00 is regardes as probe and Seq\#01 is taken as the gallery. Table \ref{comparision_oumvlp} desplays the comparison results of our method and several famous algorithms, including GEINet\cite{shiraga2016geinet}, GaitSet\cite{chao2019gaitset}, GaitPart\cite{fan2020gaitpart} and GLN\cite{hou2020gait}. It can be seen that our method can achieve the best recognition performance in most cases.

\subsection{Ablation Study} \label{Ablation_Study}
In this paper, the proposed recognition framework includes several key modules, e.g. Local Temporal Aggregation, Global and Local Feature Extractor and Generalized-Mean pooling layer. Hereby, we design different ablation studies to analyze the contribution of each key module.


\noindent \textbf{Analysis of GLFE module.}  
Different from most gait recognition methods which only extract global or local gait features, we propose a novel global and local feature extractor to extract the gait feature, which contains more comprehensive gait information. The GLFE module is composed of three global and local convolutional layers, i.e. the GLconv layers, each of which includes a global branch and a local branch. To explore the contribution of the global and local branches, we design the ablation study to explore the role of different branches. All experiments are conducted with the setting LT. The experimental results are shown in Table \ref{TAB_GLFE}. We can observe that the accuracy of using either global or local branch is 97.1\% and 96.2\% in the condition of NM, while the accuracy of using both two branches is 97.4\%, which increases by 0.3\% and 1.2\%, respectively. Meanwhile, the global and local convolution layers also improve the performance in the conditions of BG and CL. Compared with only global or local feature, the gait features extracted by GLconv can better represent the gait information. In addition, the local branch of GLconv can produce different partitions. To explore the optimal partitions in the GLConv layer, we further design the ablation study by setting different partitions. Table \ref{TAB_GLFE} indicates that the number of partitions has weak influence on accuracy. For example, the accuracy of the GLFE module with $N$=2, $N$=4, and $N$=8 is 97.3\%, 97.4\%, and 97.4\% in the condition of NM. In the method, we finally choose the GLFE module with $N$=8 to implement the model.

\begin{table}[htbp]
  \centering
  \caption{Rank-1 accuracy (\%) of different combinations in GLFE module on CASIA-B. ``GLConvA-1'' and ``GLConvA-2'' are the first and second GLConvA modules.}
  \resizebox{0.47\textwidth}{!}{
    \begin{tabular}{c|c|c|c|c|c|c|c|c}
    \toprule
    \multicolumn{2}{c|}{GLConvA-1} & \multicolumn{2}{c|}{GLConvA-2} & \multicolumn{2}{c|}{GLConvB} & \multirow{2}[2]{*}{NM} & \multirow{2}[2]{*}{BG} & \multirow{2}[2]{*}{CL} \\
\cline{1-6}    Global & N-parts & Global & N-parts & Global & N-parts &       &       &  \\
    \midrule
    \checkmark      &     $\times$  & \checkmark      & $\times$     & \checkmark      & $\times$     & 97.1  & 93.7  & 81.9  \\
    \hline
    $\times$     & 8     & $\times$     & 8     & $\times$     & 8     & 96.2  & 92.5  & 80.4  \\
    \hline
    \checkmark      & 2     & \checkmark      & 2     & \checkmark      & 2     & 97.3  & 94.3  & 82.9  \\
    \hline
    \checkmark      & 4     & \checkmark      & 4     & \checkmark      & 4     & 97.4  & 94.5  & 83.0  \\
    \hline
    \checkmark      & 8     & \checkmark      & 8     & \checkmark      & 8     & \textbf{97.4}  & \textbf{94.5}  & \textbf{83.6}  \\
    \bottomrule
    \end{tabular}%
    }
  \label{TAB_GLFE}%
  \vspace*{-1em}
\end{table}%
 
\noindent \textbf{Analysis of spatial feature mapping.} 
Traditional spatial feature mapping \cite{chao2019gaitset,fan2020gaitpart} uses $F_{Max}(\cdot)$ or $F_{Avg}(\cdot)$ to aggregate spatial information, as shown in Equ.\ref{Equ_MA}. However, using only $F_{Max}(\cdot)$, $F_{Avg}(\cdot)$, or their weighted sum, cannot realize the mapping adaptively. In this paper, we introduce the GeM pooling to implement spatial feature mapping. To verify the effectiveness of the GeM pooling layer, we design the comparison experiment by implementing methods with different spatial feature mapping strategies on the CASIA-B dataset. Note that all experiments are implemented with the LT settings. The experimental result is shown in Table \ref{comparion_spm}. 
It can be observed that the accuracy of $GeM(\cdot)$ achieves the highest average accuracy under all conditions. Specifically, the average accuracy of $F_{Max}^{1\times1\times W_{2}}(\cdot)$ and $GeM(\cdot)$ is 91.3\% and 91.8\%, respectively, meaning that the GeM pooling can obtain more robust feature representation. In addition, in the conditions of BG and CL settings, the GeM pooling performs better than all the competing pooling methods.
For example, $GeM(\cdot)$ outperforms $F_{Max}^{1\times1\times W_{2}}(\cdot)$ by 0.5\% and 1.1\% on BG and CL settings, respectively.


\begin{table}[htbp]
  \centering
  \caption{Accuracy ($\%$) of different spatial feature mapping}
  \resizebox{0.48\textwidth}{!}{
    \begin{tabular}{c|c|c|c|c|c|c}
    \toprule
    \multicolumn{3}{c|}{\textbf{Spatial Feature Mapping}} & \multirow{2}[2]{*}{NM} & \multirow{2}[2]{*}{BG} & \multirow{2}[2]{*}{CL} & \multirow{2}[2]{*}{Mean} \\
\cline{1-3}    $F_{Max}^{1\times1\times W_{2}}$   & $F_{Avg}^{1\times1\times W_{2}}$   & $GeM$   &       &       &       &  \\
    \midrule
    \checkmark      &       &       & 97.3 & 94.0  & 82.5  & 91.3  \\
    \hline
          & \checkmark      &       & 96.3  & 92.1  & 77.2  & 88.5  \\
    \hline
    \checkmark      & \checkmark      &       & 97.2  & 94.0  & 82.8  & 91.3  \\
    \hline
          &       & \checkmark        & \textbf{97.4}  & \textbf{94.5}  & \textbf{83.6} & \textbf{91.8}  \\
    \bottomrule
    \end{tabular}%
    }
  \label{comparion_spm}%
\end{table}%

\noindent \textbf{Analysis of local temporal aggregation.}
To maintain more spatial information, we introduce local temporal aggregation operation to replace the first spatial pooling layer used in traditional gait recognition methods \cite{chao2019gaitset,fan2020gaitpart}. To analyze the contribution of the LTA operation, we design the methods with different downsampling strategies. The experiments are conducted with the LT settings on the CASIA-B dataset. The experimental results are shown in Table \ref{tab_Downsampling}. It can be observed that the accuracy of using two spatial pooling layers is 96.3\%, while the accuracy of ``LTA+SP'' is 97.4\%, indicating that using the LTA operation can achieve better performance than only using spatial pooling. Besides, we also build the model by replacing both spatial pooling with LTA operation. However, the accuracy in this setting has decreased heavily. The reason may be that two LTA operations make more temporal information lost. Therefore, we take the pattern ``LTA+SP'' as the final downsampling strategy in our method.

\begin{table}[htbp]
  \centering
  \vspace*{-1em}
  \caption{Accuracy($\%$) of different combinations of downsampling}
  \resizebox{0.48\textwidth}{!}{
    \begin{tabular}{c|c|c|c|c}
    \toprule
    \multicolumn{2}{c|}{Downsampling} & \multirow{2}[2]{*}{NM} & \multirow{2}[2]{*}{BG} & \multirow{2}[2]{*}{CL} \\
\cline{1-2}    1st pooling layer & 2nd pooling layer &       &       &  \\
    \midrule
    SP    & SP    & 96.3  & 90.9  & 77.7  \\
    \hline
    SP    & LTA   & 97.0  & 93.9  & 82.2  \\
    \hline
    LTA   & SP    & \textbf{97.4} & \textbf{94.5} & \textbf{83.6} \\
    \hline
    LTA   & LTA   & 94.4  & 89.2  & 74.7  \\
    \bottomrule
    \end{tabular}%
    }
  \label{tab_Downsampling}%
  \vspace*{-1em}
\end{table}%

\section{Conclusion}
This paper proposes a novel gait recognition framework to generate discriminative feature representations under the 3D convolution formulation. 
First, to extract more comprehensive gait information, we present a global and local feature extractor to extract robust gait features for representation.
Second, to make use of more information, we also explore the effect of different downsampling approaches and introduce local temporal aggregation to replace the traditional spatial pooling layer. 
In addition, we introduce the Generalized-Mean pooling layer to adaptively aggregate the spatial information, improving the feature mapping performance. The experimental results on public datasets verify the effectiveness of the proposed method.

\noindent \textbf{Acknowledgements.}
This work was supported by the Beijing Natural Science Foundation (4202056), 
the National Natural Science Foundation of China (61976017 and 61601021), and  the Fundamental Research Funds for the Central Universities (2020JBM078). The support and resources from the Center for High Performance Computing at Beijing Jiaotong University(http://hpc.bjtu.edu.cn) are gratefully acknowledged. 

{\small
\bibliographystyle{ieee_fullname}
\bibliography{egbib}
}

\end{document}